\documentclass[conference]{IEEEtran}
\IEEEoverridecommandlockouts
\usepackage{cite}
\usepackage{url}
\usepackage{lipsum}
\usepackage[T1]{fontenc}
\usepackage{amsmath,amssymb,amsfonts}
\usepackage{algorithmic}
\usepackage{graphicx}
\usepackage{textcomp}
\usepackage{xcolor}
\usepackage{tgtermes}

\usepackage[ruled,vlined]{algorithm2e}

\def\BibTeX{{\rm B\kern-.05em{\sc i\kern-.025em b}\kern-.08em
    T\kern-.1667em\lower.7ex\hbox{E}\kern-.125emX}}
\begin{document}

\title{ PotholeGuard: A Pothole Detection Approach by Point Cloud Semantic Segmentation\\}

\author{
\IEEEauthorblockN{Sahil Nawale}
\IEEEauthorblockA{\textit{Information Technology}\\
\textit{Sardar Patel Institute of Technology}\\
Mumbai, India\\
sahil.nawale@spit.ac.in}\\   %<------ Line breaks in the current column
\IEEEauthorblockN{Gauransh Sawhney}
\IEEEauthorblockA{\textit{Dept. of Electrical Electronics}\\
\textit{BITS Pilani}\\
Pilani, India\\
f20180325@pilani.bits-pilani.ac.in}
\and
\IEEEauthorblockN{Dhruv Khut}
\IEEEauthorblockA{\textit{Information Technology}\\
\textit{Sardar Patel Institute of Technology}\\
Mumbai, India\\
dhruv.khut@spit.ac.in}\\   %<------ Line breaks in the current column
\IEEEauthorblockN{Pushkar Aggrawal}
\IEEEauthorblockA{\textit{Dept. of Electrical Electronics}\\
\textit{BITS Pilani}\\
Pilani, India\\
f20180431@pilani.bits-pilani.ac.in}
\and
\IEEEauthorblockN{Daksh Dave}
\IEEEauthorblockA{\textit{Dept. of Electrical Electronics}\\
\textit{BITS Pilani}\\
Pilani, India\\
f20180391@pilani.bits-pilani.ac.in}\\
\IEEEauthorblockN{Dr. Kailas Devadkar}
\IEEEauthorblockA{\textit{Dept. of Computer Science and Engineering}\\
\textit{Sardar Patel Institute of Technology}\\
Mumbai, India\\
kailas\_devadkar@spit.ac.in}\\   %<------ Line breaks in the current column
}

\maketitle

\begin{abstract}
Pothole detection is crucial for road safety and maintenance, traditionally relying on 2D image segmentation. However, existing 3D Semantic Pothole Segmentation research often overlooks point cloud sparsity, leading to suboptimal local feature capture and segmentation accuracy. Our research presents an innovative point cloud-based pothole segmentation architecture. Our model efficiently identifies hidden features and uses a feedback mechanism to enhance local characteristics, improving feature presentation. We introduce a local relationship learning module to understand local shape relationships, enhancing structural insights. Additionally, we propose a lightweight adaptive structure for refining local point features using the K-nearest neighbor algorithm, addressing point cloud density differences and domain selection. Shared MLP Pooling is integrated to learn deep aggregation features, facilitating semantic data exploration and segmentation guidance. Extensive experiments on three public datasets confirm PotholeGuard’s superior performance over state-of-the-art methods. Our approach offers a promising solution for robust and accurate 3D pothole segmentation, with applications in road maintenance and safety.
\end{abstract}


\section{Introduction}

Pothole identification is a critical issue in road maintenance and safety. Research has shown that repair costs for damaged pavements can increase sevenfold in just five years \cite{wang20233d}. Potholes also contribute to vehicle suspension damage, compromising ride comfort and increasing accident risks. Regular inspection and prompt pothole repair are essential for road maintenance. Well-maintained roads also play a crucial role in reducing the likelihood and severity of accidents. As a result, the demand for automated pothole detection systems is growing.

Deep learning methods excel segmentation due to their ability to automatically learn features, handle non-linearity, adapt to varying point densities, capture hierarchical structures, and provide robustness to noisy data. Their scalability and  performance make them a popular choice over other Machine Learning algorithms.
Image segmentation for potholes typically relies on two-dimensional data and spatial attributes, assessing severity. Segmenting 3D point cloud data \cite{cheng2021sspc} can enhance the feature extraction and pothole detection process for pothole detection\cite{fan2019pothole}.
However point cloud segmentation poses many difficulties like  large segmentation time and resources required\cite{xu2021rpvnet}, loss of geometric details and inability to capture local intricate feature effectively. Thus semantic segmentation of 3D point clouds remains challenging \cite{fan2021scf}.

To address these challenges and extract local-global contextual features, we propose an efficient and accurate model inspired by SCF-Net \cite{fan2021scf}. Our approach enriches semantic information using a feedback mechanism. Segmentation over other techniques provides exact structure of the pothole rather than simple detection, faciliating further work like prioritizing potholes based on their severity, or estimating the actual interal condition of the road. Our contributions are as follows:

\begin{itemize}
    \item A proposed module offering advanced capabilities for local shape relation analysis and structural detail extraction within point cloud data of varying densities, enabling precise detection of potholes even in challenging scenarios.
    \item Introduced a Feature Augmenter which plays a critical role in uncovering latent data features, explicitly encoding intricate relationships between local and global features. This results in substantial enhancement of the original local feature set, improving the representation of geometric attributes in point cloud data.
    \item Extensive comparison with traditional and SOTA (State-Of-The-Art) methods.
\end{itemize}

\section{Related Works}

Earliest methods for pothole detection used 2D image segmentation \cite{akagic2017pothole}.
However accuracy of 2D methods drops drastically in many cases, like poor illumination and visibility. Furthermore, since 2D methods lack a sense of depth, they cannot extract the spatial geometric structure and volume of the pothole. Hence a shift to 3D methods was necessary. Most of the existing 3D methods employ 3D CNN's \cite{wang20233d} for point cloud segmentation. Some attempt to adapt 2D segmentation methods by projecting 3D point clouds into 2D images and then applying 2D semantic segmentation techniques. However, this projection inevitably results in the loss of detailed information.

% To capture contextual features and geometric structures, some works have explored segmentation using graph networks \cite{wang2019graph} and Recurrent Neural Networks (RNN's) \cite{zhang2019automated}. EdgeConv \cite{wang2019dynamic}, a lightweight convolution kernel, has demonstrated exceptional performance in point cloud classification and segmentation. DGCNN \cite{xiong2022dcgnn}, semantically groups points by dynamically updating a relationship graph across layers, using the EdgeConv operator to better capture local features. 

Discretization-based methods convert point clouds into discrete representations \cite{choy2019fully}, such as voxels, which are 3D grids, and processed using fully-3D CNNs for voxel-wise segmentation. These are adept at handling large point clouds, but their performance depends on voxel granularity, often introducing discretization artifacts. Hence PointNet\cite{DBLP:journals/corr/QiSMG16} was introduced, which applies MLPs to each point and aggregates individual point features into global features through max-pooling (MP) but cannot effectively capture local features, which are vital for point cloud analysis.
PointNet++\cite{DBLP:journals/corr/QiYSG17} addressed this issue, treating PointNet as a local feature learner in a hierarchical architecture, capturing local features through farthest-point sampling (FPS) and k-nearest neighbors (KNN). 

Other approaches also contribute to the field \cite{guo2021pct,lai2022stratified}, but most of them suffer from some major or minor problems. For instance, Super Point Graph (SPG)  \cite{sarlin2020superglue} and PCT are effective at capturing most of the features of the point cloud, but incur heavy overhead thereby decreasing efficiency. In the case of PCT the reason for low efficiency is the inherent complex and bulky nature of the transformers, requiring high computation power and large processing time even for basic point clouds, while in the case of SPG, it is because of the requirement of preprocessing of point clouds into super point graphs and heavy dependence on geometrically homogeneous partitioning, which sometimes even may lead to incorrect segmentation. 

To efficiently extract point cloud features, RandLA-Net \cite{hu2020randla} was introduced, which achieves high efficiency through random sampling and employs a local feature aggregation module to learn and preserve geometric patterns. However, it is able to do so only in the case of smaller and relatively simple point clouds. 
Similarly, SCF-Net\cite{fan2021scf} is another efficient approach, but its standalone SCF local feature extractor may not effectively capture complex structures, resulting in reduced capability while extracting intricate point cloud features.

Unlike the methods, the proposed method is fast because it doesnt involve reconstruction of point cloud from images. Our method is accurate as it able to detect most of the hidden features effectively. Furthermore segmentation over simple detection helps detailed analysis of exact structure and condition of the pothole and the road, allowing to estimate the severity of damage and cost of repairs. Finally, the network is highly flexible and allows any other modules to be easily integrated in the network.

\section{Methodology}
\label{sec:methodology}

\begin{figure*}[h!]
   \centering
   \includegraphics[width=\linewidth]{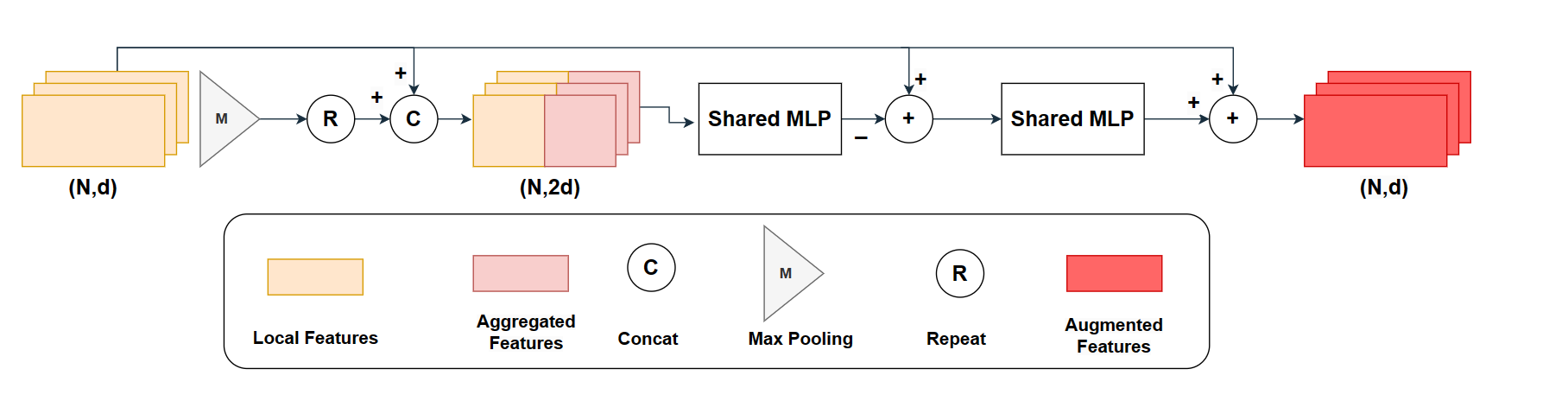}
   \caption{Feature Augmenter Module}
   \label{fig:cclogo1}
\end{figure*}

This section focuses on PotholeGuard's architectural design. Technologies like Lidar sensors and photogrammetry are used to construct the pothole point clouds. The created set of input points is represented as P, given by 

\begin{equation}
P=\left\{p_1, p_2, p_3 \ldots p_i \mid 1 \leq i \leq N, i \in \mathbb{N}\right\} \\
\end{equation}

\begin{equation}
p_i=\{x, y, z . . d\} \in \mathbb{R}^d
\end{equation}

where N is number of points in the point cloud, x,y,z...d are the dimensions/features of each individual point in the point cloud. The points are first directed through the feature augmenter module, simplifying the process of feature extraction from the point cloud for the model.

\vspace{3mm}
\subsubsection{Feature Augmenter}

The Feature Augmenter module represented in Figure 1 serves a dual role within the model. Firstly, it acts as an initial preprocessing step, enhancing the features to facilitate easier feature extraction by the model. For example, the surface deformations and texture variations can be augmented in the beginning to mark a clear distinction between the surface of the road and the pothole. Secondly, it is strategically positioned at the bottleneck region of SCF-Net where it specializes in learning intricate and visually obscure local and global features as well as augmenting the features of high-dimensionality points within the encoder, complementing the initial features that tend to diminish as they traverse the network. The Feature augmentation is achieved by repeated max-pooling on the local features. The local features are concatenated with features from max-pooling to retain the basic features such as the background. Finally these concatenated features are blended using MLP.

The N points in the Point Cloud with d features are passed into the network. Initially, the local features of these points are converted into global features g using a max pooling function h():
\begin{equation}
g=h\left(p_1, p_2, \ldots, p_N\right) \in \mathbb{R}^{1 \times d}
\end{equation}

The global feature g is repeated N times and concatenated with original local features to generate local-global aggregated features.
This can be given by the equation : 

\begin{equation}
\bar{p}_i=\operatorname{MLP}\left(p_i \oplus \operatorname{repeat}(g)\right) \in \mathbb{R}^{N \times d}
\end{equation}

where $\bar{p}_i$  represents the local-global aggregated features.

While our approach shares a conceptual similarity with auto-encoders \cite{cui2020lightweight}, there are distinctive differences between them. 
Auto-encoders typically modify the size of the point cloud and aggregate features from various resolutions to achieve a comprehensive representation, 
however we integrate the local features and the local-global aggregated features with cascaded series repetition encoding via group of MLP's which directly  result in generation of the augmented features all while maintaining the number of points in the point cloud and changing only the number of features/dimensions of these points.

\begin{equation}
\bar{p}_{f_i}=p_i+\operatorname{MLP}\left(\bar{p}_i-p_i\right) \in \mathbb{R}^{N \times d}
\end{equation}

where $\bar{p}_{f_i}$ is the individual point in the point cloud with augmented features. The module is integrated in a serial cascaded manner in SCF-Net.

\vspace{3mm}

\subsubsection{PotholeGuard Module}
\begin{figure*}[h!]
   \centering
   \includegraphics[width=\linewidth]{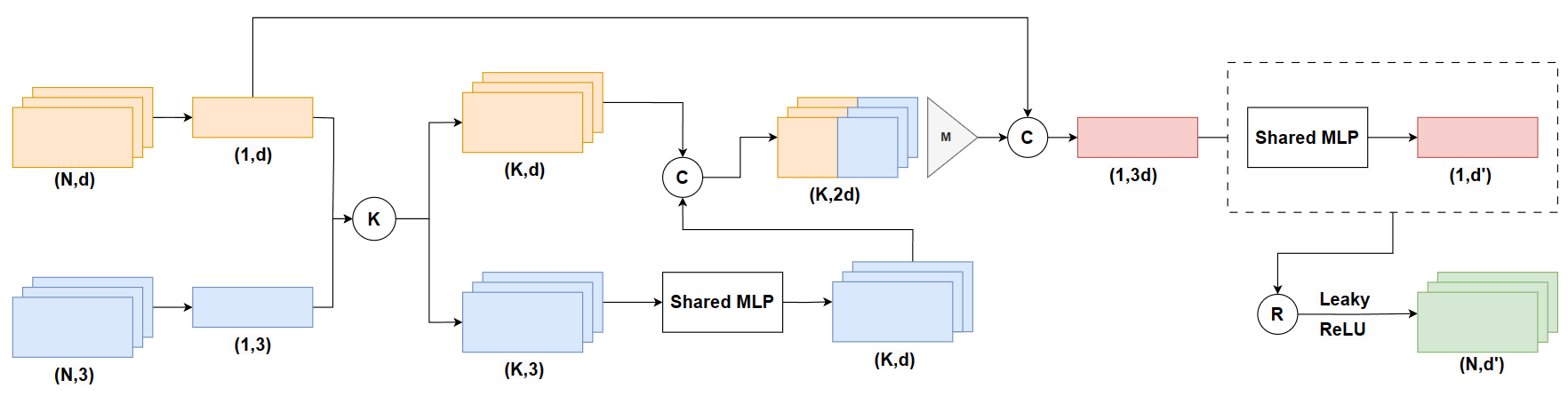}
   \caption{PotholeGuard Module}
   \label{fig:cclogo2}
\end{figure*}

The Module in Figure 2 receives input comprising spatial information and augmented features derived from the initial Feature Augmenter. Spatial information provides insights into  local and global contextual features, while the augmented features serve as a substrate for the acquisition of nuanced, hidden local features, like subtle depth variations and determinination of the exact boundary of the pothole. The local contextual features undergo a dual concatenation process to enrich the contextual information. Subsequently, these augmented features are subjected to an additional operation of summation with another feature map, culminating in the refinement of the local feature set.

The characteristics pertaining to the distribution of neighboring points primarily encapsulate the spatial structure of pothole and the distance between a point and the central point of its local neighbourhood. In cases where the neighboring points exhibit even distribution, the center point will closely align with the given point. Let $p_i$ be a point in the point cloud, and $\left\{p_j\right\}$ the set of neighboring points to it, where j=0,1, ...k, determined through the K-Nearest Neighbors algorithm. Consequently, the distribution characteristics, $L_i$, can be expressed as : 

\begin{equation}
L_i=\left\|p_i-\frac{\sum_{j=0}^k p_j}{k}\right\|
\end{equation}

with ||.|| signifying the centroid. The coordinates of neighbor points relative to the centroid, the centroid's coordinates, and the distances from the neighbor points to the centroid, are amalgamated. The module integrates this information with the features of the neighboring points as :

\begin{equation}
\left\{\hat{p_{i j}}\right\}=\left[\left\{r_{i j}\right\},\left\{p_{i j}\right\}\right]
\end{equation}

Then $\hat{p_{i j}}$ is passed through a series of mapping functions using repeated MLP after passing it through MaxPooling to extract the salient features of irregular points in Point Cloud of Pothole. Unlike PointNet++, the latter part of our Module can be repeated number of times to increase the depth, trainable parameters, adaptability and accuracy of the model for a trade-off with increased resources. The specific expression of this part can be given by : 
\begin{equation}
g_{i j}=A\left(\Phi\left(\hat{p_{i j}}\right)\right)
\end{equation}
\begin{equation}
\hat{g}_{i j}=\operatorname{ReLU}\left(R\left(M L P\left(g_{i j}\right)\right)\right)
\end{equation}
    where $A$ is MaxPooling and $\phi$ is the sequence of mapping operations, R is repetition $\hat{g}_{i j}$ is the final processed points of the point cloud with the required number of output features. The resulting output of this module consists of the learned spatial contextual feature, a composite construct that seamlessly integrates both local and global contextual attributes.

\vspace{3mm}
\subsubsection{Main Network}

\begin{figure*}[h!]
   \centering
   \includegraphics[width=\linewidth]{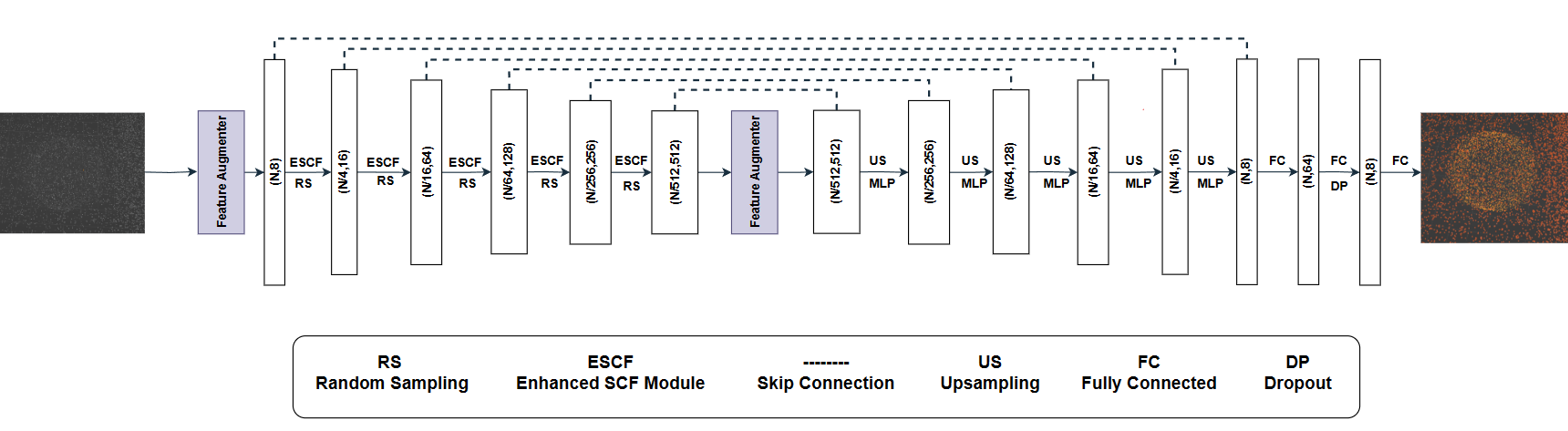}
   \caption{Architecture of the Network}
   \label{fig:cclogo3}
\end{figure*}

Figure 3 depicts the main network architecture, built upon a standard encoder-decoder framework. Initially, N input points traverse the Feature Augmenter module, refining feature accuracy. Subsequently, five encoder layers systematically encode features, employing random sampling for dimension reduction, and the SCF module for contextual feature learning. This reduces points from N to N/512, with the feature dimension growing from 8 to 512. Another Feature Augmenter at the bottleneck maintains feature consistency, vital for up-sampling and interpolation. Five decoder layers decode features, undergoing up-sampling via nearest-neighbor interpolation and concatenation with intermediate features through skip connections. Semantic labels are predicted through three fully connected layers, with dropout enhancing generalization and robustness in the network's predictions.

\section{Experimental Results}
\label{sec:results}
\begin{table*}[]
\caption{Results on S3DIS Dataset}
\begin{tabular}{|l|l|l|l|l|l|l|l|l|l|l|ll}
\cline{1-11}
\textbf{Methods}           & \textbf{OA(\%)} & \textbf{mAcc(\%)} & \textbf{mIoU(\%)} & Ceiling       & Flooring      & Wall          & Beam          & Column & Window        & Door          &  &  \\ \cline{1-11}
PointNet\cite{DBLP:journals/corr/QiSMG16}                   & 78.6            & 66.1              & 47.4              & 88.8          & 97.3          & 69.8          & 42.4          & 23.1   & 46.3          & 51.6          &  &  \\ \cline{1-11}
PointNet++\cite{DBLP:journals/corr/QiYSG17}                 & 80.5            & 66.9              & 53.5              & -             & -             & -             & -             & -      & -             & -             &  &  \\ \cline{1-11}
KPConv\cite{thomas2019kpconv}                     & -               & 72.8              & 67.1              & 93.6          & 92.4          & 81.2          & 63.9          & 45.1   & \textbf{66.1} & 69.2          &  &  \\ \cline{1-11}
RandLA-Net\cite{hu2020randla}                 & 88              & \textbf{82}       & 70                & 93.1          & 96.1          & 80.6          & 62.4          & 48     & 64.4          & 70.1          &  &  \\ \cline{1-11}
SCF-Net\cite{fan2021scf}                    & 71.6            & 76.2              & 69.8              & 93.3          & 96.4          & 80.9          & 64.9          & 47.4   & 64.5          & \textbf{70.1} &  &  \\ \cline{1-11}
Point Transformer\cite{guo2021pct}          & 90.8            & 76.5              & 70.4              & 95.2          & 96.8          & 81.3          & 64.9          & 45.6   & 64.4          & 66.3          &  &  \\ \cline{1-11}
Lai Stratified Transformer\cite{lai2022stratified} & 91.5            & 78.1              & 72.0              & 95.5          & \textbf{98.2} & \textbf{82.9} & 65.5          & 45.6   & 65.7          & 67.0          &  &  \\ \cline{1-11}
FA-ResNet\cite{zhan2023fa}                  & 89.0            & 76.9              & 68.1              & 94.0          & 97.9          & 82.3          & 65.3          & 45.1   & 64.1          & 65.2          &  &  \\ \cline{1-11}
\textbf{PotholeGuard}              & \textbf{91.5}   & 77.2              & \textbf{75.3}     & \textbf{95.9} & 98.1          & 82.2          & \textbf{65.7} & 45.8   & 65.9          & 67.7          &  &  \\ \cline{1-11}
\end{tabular}
\end{table*}
\subsection{Experimental Setup}
Our setup uses a batch size of 1 and initial learning rate of 0.02, with decay rate of 0.95 over 100 training epochs, optimized using the Adam optimizer. The KNN algorithm is configured with a neighborhood size of 16. The model is trained on a mix of S3DIS and a pothole point cloud dataset\cite{fan2019pothole}, which contains 67 labeled point clouds. The model has been tested on more 80 point clouds constructed using iPhone 14's Lidar sensor and photogrammetry. Our system has a NVIDIA GeForce GTX 1650 GPU, an AMD Ryzen 5 4600H Processor and 16GB of RAM. The code is executed within a Python virtual environment, using VSCode code editor, and on an Amazon AWS EC2 instance for scalable performance. TensorFlow serves as the core modeling framework, complemented by auxiliary libraries such as NumPy, SciPy, and h5py to support various aspects of the research. The point clouds are visualized in Blender software because of its ability to simulate large point clouds even on low end devices.

\subsection{Segmentation Results}
Due to the scarcity of publicly available pothole-specific datasets and the prevalence of methods applied to reconstructed point clouds from 2D images, direct method-to-method comparisons are challenging. Hence, we conduct a comprehensive evaluation of our model's performance, encompassing both general segmentation tasks and rigorous testing on established public point cloud datasets, as elaborated in the upcoming section.  
For the evaluation metrics, we use mean of class-wise intersection over union (mIoU), mean
of class-wise accuracy (mAcc), and overall point-wise accuracy (OA). 
Segmentation results of the model on point cloud constructed using iPhone Lidar is showcased in Figures 4, 5, and 6. Figure 4 offers a top-view representation of the segmentation results for a manhole, while Figures 5 and 6 underscore the effectiveness of our segmentation methodology in accurately detecting potholes on road surfaces.
\begin{figure}
    \centering   \includegraphics[width=\linewidth]{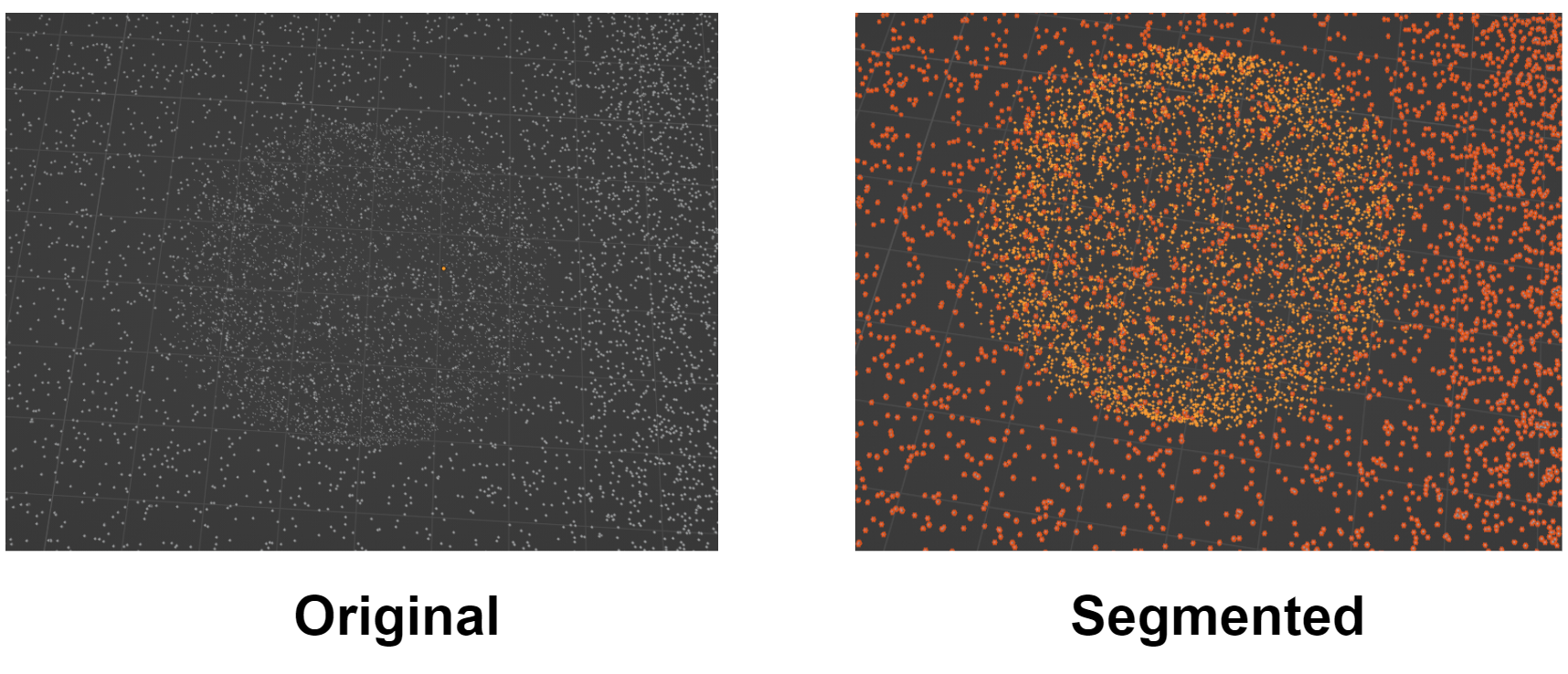}
   \caption{Segmentation Results for a Pothole with Dense Point Populations}
   \label{fig:cclogo4}
\end{figure}
% \begin{figure}[]
%    \centering
%    \includegraphics[width=\linewidth]{res2.png}
%    \caption{Segregation Outcome on Sparse Point Populations within a Pothole}
%    \label{fig:cclogo5}
% \end{figure}

\begin{figure}
   \centering
   \includegraphics[width=\linewidth]{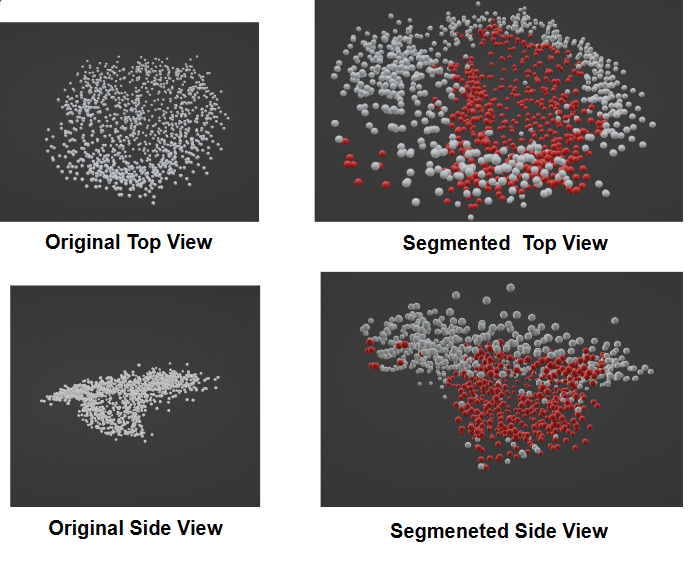}
   \caption{Pothole Segmentation Result with Sparse Points in Multiple Views}
   \label{fig:cclogo6}
\end{figure}
\subsection{Analysis}
The following section reports comparison results of PotholeGuard on two public datasets, S3DIS \cite{armeni20163d} and ScanObjectNN \cite{uy-scanobjectnn-iccv19} datasets. 

\begin{itemize}
\item ScanObjectNN : ScanObjectNN is a collection of point cloud of real-world objects with about 15k objects in 15 different categories. The point clouds in the dataset are much more difficult to segment and classify due to the presence of irregular density point clouds and partial occlusions.
\item S3DIS : This dataset was issued in 2016. It contains 6 large-scale indoor areas with 271 rooms. Each point in the scene point cloud is annotated with one of the 13 semantic categories.
\end{itemize}

Table 1 presents a comprehensive comparison of our architecture with other state-of-the-art methods. Highest-performing metrics are highlighted in bold. 
Underperformance of the network in some other metrics is justified as the model has been trained on a mix of pothole pointcloud and general object's pointcloud from S3DIS dataset as well as insufficient training due to lack of computational power. Nevertheless, the proposed architecture excels across various comparison metrics, achieving the highest Overall Accuracy (91.5\%) and Mean Intersection over Union (mIoU), a metric that evaluates the accuracy of pixel-wise object segmentation, with a value of 75.3\%. Furthermore, it demonstrates remarkable performance in individual categories, notably Ceiling (95.9\%) and Beam (65.7\%). Table 2 showcases our method's exceptional performance on the ScanObjectNN dataset. Compared to other CNN-based methods and PointNet, our approach stands out with an impressive Overall Accuracy (OA) of 82.3\% and a remarkable mean accuracy (mAcc) of 81.6\%. These results underscore the effectiveness and superiority of our architecture.

\begin{table}[]
\centering
\caption{Results on ScanObjectNN Dataset}
\begin{tabular}{|l|l|l|l|ll}
\hline
\textbf{Method}         & \textbf{mAcc(\%)}      & \textbf{OA(\%)}        \\ \hline
PointNet++\cite{DBLP:journals/corr/QiYSG17}     & 75.4          & 77.9          \\ \hline
BGA-PointNet++\cite{uy2019revisiting} & 77.5          & 80.2          \\ \hline
DGCNN\cite{xiong2022dcgnn}         & 73.6          & 78.1          \\ \hline
AdaptConv\cite{zhou2021adaptive}      & 76            & 79.3          \\ \hline
PotholeGuard           & \textbf{81.6} & \textbf{82.3} \\ \hline
\end{tabular}
\end{table}

Table 3 presents a comparison of the number of trainable parameters and its corresponding segmentation time. Notably, our model has large number of trainable parameters, about 15.225 million, contributing to its remarkable accuracy,but at the cost of slightly longer segmentation time, approximately 30 seconds more than other models of comparable accuracy. However, it still offers significantly improved speed compared to transformer-based models with nearly identical levels of accuracy. Compared to traditional 2D pothole image segmentation methods, like the Unet++\cite{zhou2018unet} which has segmentation time of 55ms, 3D segmentation is able to draw out much more information than 2D segmentation at mere runtime of 140ms. However, 2D methods are easy to use and integrate since no special 3D simulation softwares are required. This balance between parameter complexity and processing time underscores the practical efficiency of our proposed approach.

\begin{table}[]
\centering
\caption{Parameter Count and Time Comparison Among Methods}
\begin{tabular}{|l|l|l|ll}
\hline
           & \textbf{Parameters(millions)} & \textbf{Total Time(seconds)} \\ \hline
RandLA-Net\cite{hu2020randla} & 4.993                & 114.89              \\ \hline
SCF-Net\cite{fan2021scf}    & 12.146               & 119.39              \\ \hline
SCFL-Net\cite{liu2023ss}   & 12.147               & 119.90              \\ \hline
PotholeGuard       & 15.225               & 140.67              \\ \hline
\end{tabular}
\end{table}

\begin{figure}[h!]
   \centering
   \includegraphics[width=\linewidth]{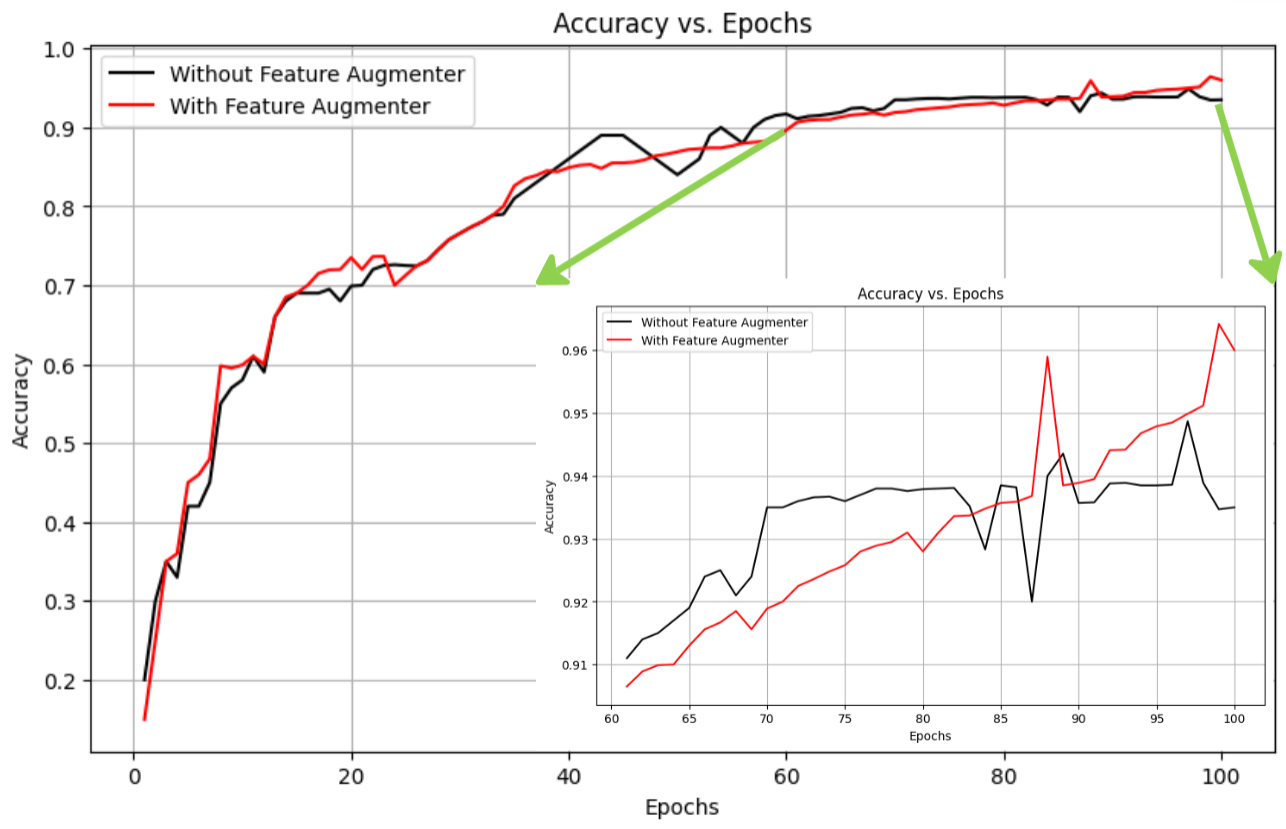}
   \caption{Results with and without Feature Augmenter module}
   \label{fig:cclogo7}
\end{figure}

Figure 6 shows the training accuracy versus number of epochs with and without the Feature Augmenter module. The graph reveals that including the module model proves to be instrumental for fine-tuning in the later stages of training, thereby increasing accuracy, because the network is able to detect even more intricate features as the training proceeds, as these intricate features are augemented by the Feature augmenter module, enabling them to be captured effectively. Furthermore it provides the network with a greater stability while training, as a result of larger number of trainable parameters, contributed by the layers in Feature Augmenter. The network with less number of parameters usually overfits or underfits to specific features in specific epochs causing fluctuations, unlike the network with larger number of trainable parameters which can encompass wide range of features due to its larger non-linearity.

\section{Conclusion}
\label{sec:conclusion}
In this paper, we have introduced PotholeGuard, a pioneering model for 3D Point Cloud segmentation. Within this framework, we have unveiled two novel modules: the PotholeGuard Module and the Feature Augmenter, strategically integrated into the core SCF Network to enhance the model's performance. These modules facilitate the precise capture of intricate local-global contextual features, elevating their representation through augmentation. Our extensive experimentation has yielded compelling results in Pothole Point Cloud Segmentation. Notably, PotholeGuard achieves an overall accuracy (OA) that is nearly on par with the state-of-the-art Lai Stratified Transformer, while surpassing it by approximately 1.3\% in mean Intersection over Union (mIoU). Furthermore, it outperforms competing models on the ScanObjectNN dataset, achieving an approximately 3.7\% increase in mean accuracy (mAcc) and a 1.3\% boost in overall accuracy (OA) over the second-best performing models. These findings underscore PotholeGuard's efficacy in addressing complex point cloud segmentation tasks, highlighting its potential for real-world applications such as real-time detection of potholes to avoid incidents. This automation of pothole detection saves authorities large time and resources and faciliate assessing the severity of the pothole or damage done to the road.

% \bibliographystyle{plain}
% \bibliography{my_bib}

% \addbibresource{my_bib.bib}
\bibliographystyle{IEEEtran}
\bibliography{my_bib}

\end{document}